\documentclass[cameraready]{Interspeech}

\newcommand{\bcirc}[1]{\raisebox{0.5pt}{\textcircled{\footnotesize\bfseries #1}}}
\usepackage{comment}
\usepackage{pifont}
\usepackage{algorithm}
\usepackage{algpseudocode}
\usepackage{amsmath,amssymb}
\usepackage{graphicx}  
\usepackage[table]{xcolor}
\usepackage{colortbl}
\usepackage{booktabs}
\usepackage{adjustbox}
\usepackage{multirow}
\usepackage{array}
\usepackage{subcaption}
\usepackage{float}
\title{Selective Capability Unlearning in End-to-End Spoken Language Understanding}

\author[affiliation={1}, orcid=0009-0005-1624-272X]{Akanksha}{Singh}
\author[affiliation={1}, orcid=0000-0003-0332-5647]{Vinod Kumar}{Kurmi}

\address{
$^1$ Indian Institute of Science Education and Research Bhopal, India
}

\email{akanksha23@iiserb.ac.in, vinodkk@iiserb.ac.in}

\usepackage{comment}

\begin{document}

\maketitle

\begin{abstract}
Modern spoken language understanding (SLU) systems are increasingly deployed in real-world settings, where specific functionalities may need to be removed due to policy or safety constraints. In SLU, a functionality corresponds to an intent and its associated slot-generation behavior. However, in autoregressive models, suppressing a target intent does not eliminate the conditional mapping that generates slots conditioned on that intent. When the intent prefix is externally supplied, the model can reconstruct the original intent-slot structure. We identify this structural failure as \textbf{\emph{capability persistence}}. We propose \textit{\underline{B}inding \underline{S}ubspace (BSU)}, a representation-level framework that isolates and attenuates intent-conditioned directions underlying this mapping. Across SLU benchmarks, BSU substantially reduces forced-prefix recoverability while preserving retained performance.
\end{abstract}

\section{Introduction}
Spoken language understanding (SLU) constitutes a core component of conversational systems. It enables devices like voice assistants and spoken interfaces to extract structured semantic information directly from speech \cite{bastianelli2020slurp, wang2021fine, arora2022espnet, seo2022integration}. Modern end-to-end SLU models \cite{huang2023leveraging, sharma2021leveraging} directly map acoustic input to semantic outputs and are widely adapted in diverse applications, including virtual assistants, customer support systems, in-vehicle voice control, and domain-specific voice agents \cite{ma2024speech, mehta2020recent}.

\vspace{0.2em}
As these systems are increasingly deployed in real-world and regulated settings, there is a growing need to adjust model behavior post-training, including SLU frameworks \cite{koudounas2025alexa}. In practice, certain functionalities may need to be disabled due to policy changes, safety considerations, or regulatory requirements \cite{voigt2017eu, goldman2020introduction, xu2024machine}. For example, a voice assistant may need to deactivate financial transaction capabilities in specific regions or restrict health-related guidance under updated compliance rules. Retraining these models from scratch in such cases is often costly and impractical, motivating the need for targeted post-deployment methods that can selectively remove undesired behaviors while preserving other functionalities.

\begin{figure}[t]
  \centering
  \includegraphics[width=\columnwidth]{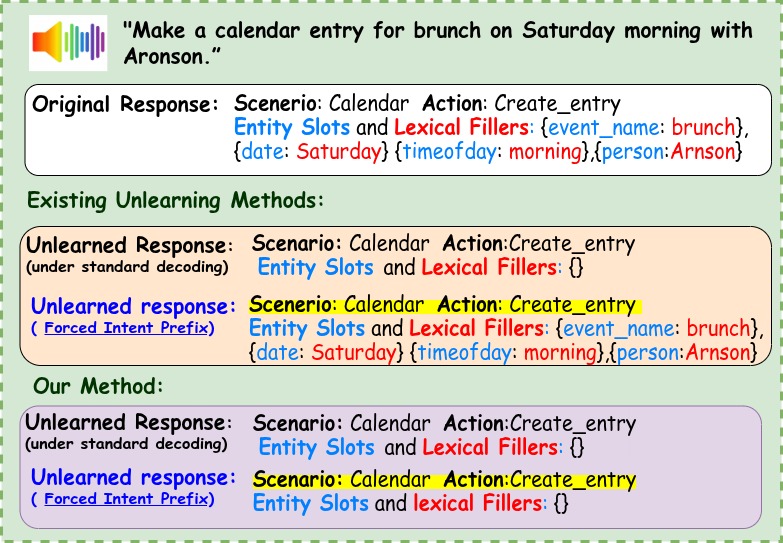}
  \vspace{-0.45cm}
    \caption{\textbf{Capability persistence under forced-prefix decoding.} Existing methods suppress marginal intent prediction but preserve conditional intent–slot binding, enabling reconstruction under forced prefix. BSU removes this dependency.}
  \label{fig:01}
\end{figure}

\vspace{0.2em}
However, in an autoregressive SLU, functionalities are not merely intent labels. They correspond to an intent together with its associated slot-generation behavior. Semantic generation is inherently conditional, as the decoder first predicts an intent token, and subsequent slot tokens are generated conditioned on that prefix. Thus, the behavior associated with an intent is governed by the conditional mapping from acoustic input and intent prefix to slot values. In this work, we define the removable functionality as precisely this intent-conditioned conditional mapping. Upon a request to remove functionality, existing methods typically suppress the marginal probability of a target intent. However, this does not necessarily modify the conditional mapping responsible for slot generation. Consequently, under a forced intent prefix, the model can still reconstruct the corresponding intent-slot structure. We refer to this phenomenon as capability persistence. As illustrated in Figure~\ref{fig:01}, although intent prediction is suppressed under standard decoding, providing the intent as a prefix enables recovery of the associated slot structure. This limitation arises from the conditional generative structure of autoregressive SLU models, which classification-based unlearning methods do not explicitly address. To eliminate the conditional mapping itself, we propose a two-stage framework.

\vspace{0.3em}
\noindent\bcirc{1}\ \textbf{Binding Subspace Identification.}
For a target intent $I_F$, we identify representation directions that capture its slot-generation behavior. We extract decoder hidden states at slot positions for forget and retain data and compute layer-wise covariances. By contrasting these covariances, we extract the eigen-directions with the largest positive variance excess under $I_F$, forming a compact subspace that captures intent-slot dependency.

\noindent\bcirc{2}\ \textbf{Subspace-Guided Capability Attenuation.}
We fine-tune the model while reducing sensitivity along the identified subspace. Specifically, we apply gradient-based regularization to attenuate dependence on these directions, thereby weakening the conditional mapping while preserving general performance.

\vspace{0.3em}
Our key contributions are outlined as follows:
\textbf{\textit{(i)}} We formalize selective capability unlearning in SLU by defining intent capability as the conditional mapping and identifying \emph{capability persistence} as recoverable slot generation under forced-prefix decoding.
\textbf{\textit{(ii)}} We propose \textit{Binding Subspace Unlearning} (BSU)\footnote{Annotations and code will be made publicly available.}, a representation-level framework that localizes intent-conditioned binding directions via covariance contrast and attenuates them through subspace-guided gradient regularization.
\textbf{\textit{(iii)}} We introduce a recoverability-based evaluation protocol to measure residual conditional behavior beyond intent accuracy.
\textbf{\textit{(iv)}} BSU reduces conditional slot recoverability, with an average drop of $\sim$60\% in BRR@10 and $\sim$56\% in semantic similarity, while preserving retained-intent performance without inference-time overhead.

\section{Problem Formulation}
\vspace{-0.1cm}
We consider an end-to-end spoken language understanding (SLU) model that maps an input speech signal $x \in \mathcal{X}$ to a structured semantic frame represented as a token sequence $y=(y_1,\dots,y_T)$. The sequence follows a fixed format in which the initial tokens encode an intent label $i \in \mathcal{I}$, followed by slot-type and slot-value tokens $s \in \mathcal{S}$. The model is autoregressive and parameterized by $\theta$, defining
\begingroup
\setlength{\abovedisplayskip}{3pt}
\setlength{\belowdisplayskip}{3pt}
\setlength{\abovedisplayshortskip}{3pt}
\setlength{\belowdisplayshortskip}{3pt}
\[
p_{\theta}(y \mid x)
=
\prod_{t=1}^{T}
p_{\theta}(y_t \mid y_{<t},x).
\]
\endgroup
Because intent tokens precede slot tokens in the decoding order, the joint distribution factorizes as
\begingroup
\setlength{\abovedisplayskip}{4pt}
\setlength{\belowdisplayskip}{4pt}
\setlength{\abovedisplayshortskip}{4pt}
\setlength{\belowdisplayshortskip}{4pt}
\[
p_{\theta}(i,s|x)
=
p_{\theta}(i|x)\,
p_{\theta}(s|i,x),
\]
\endgroup
where slot generation is conditioned jointly on the acoustic input and the decoded intent prefix. We define the semantic capability associated with intent $i$ as the conditional mapping
\begingroup
\setlength{\abovedisplayskip}{4pt}
\setlength{\belowdisplayskip}{4pt}
\setlength{\abovedisplayshortskip}{4pt}
\setlength{\belowdisplayshortskip}{4pt}
\[
C_{\theta}(i)(x)=p_{\theta}(s|i,x),
\]
\endgroup
that characterizes the model’s ability to generate slot values given the intent and acoustic input. This definition distinguishes capability from marginal intent prediction. Suppressing the marginal probability $p_{\theta}(i|x)$ does not necessarily eliminate the conditional mapping $p_{\theta}(s|i,x)$ when the intent prefix is externally supplied during decoding. This motivates the problem of \textit{selective capability unlearning}.
\vspace{-1em}
\subsection{Selective Capability Unlearning } 
\vspace{-0.5em}
Let $I_F \in \mathcal{I}$ denote a target intent whose associated capability must be removed. The training corpus is partitioned into a forget set $\mathcal{D}_F$, containing samples labeled with $I_F$, and a retain set $\mathcal{D}_R$, containing all remaining intents. Given pretrained parameters $\theta_0$, the objective is to obtain parameters $\theta^*$ such that:
\begin{enumerate}[label=\ding{\numexpr171+\arabic*}, leftmargin=2em]
\item \textbf{Capability Erasure}. The conditional mapping $p_{\theta^*}(s \mid i_f, x)$ must be substantially degraded for inputs $x \sim \mathcal{D}_F$, such that even when the intent prefix $I_F$ is externally supplied, the model can no longer generate correct slots.
\item \textbf{Capability Retention}. For all $i \neq i_f$, the joint behavior $p_{\theta^*}(i, s \mid x)$ should remain close to that under $\theta_0$, thereby preserving performance on all retained intents.
\end{enumerate}
In other words, our objective is to reduce the conditional mapping associated with the target intent while preserving the remaining capabilities of the model.

\vspace{-0.1cm}
\subsection{Failure Mode: Capability Persistence} 
Existing unlearning methods typically suppress the marginal likelihood $p_{\theta}(i_f \mid x)$ of the target intent. However, this does not 
necessarily modify the conditional slot-generation distribution 
$p_{\theta}(s \mid i_f, x)$. This limitation arises from the autoregressive decoding. For slot positions 
$t > t_{\text{intent}}$, the decoder hidden state $h_t = f_{\theta}(x, i_f, s_{<t})$
depends jointly on the acoustic input $x$, the intent prefix $I_F$, and 
previously generated tokens. The next-token probability is computed as
\begingroup
\setlength{\abovedisplayskip}{1pt}
\setlength{\belowdisplayskip}{2pt}
\setlength{\abovedisplayshortskip}{1pt}
\setlength{\belowdisplayshortskip}{2pt}
\[
p_{\theta}(y_t \mid y_{<t}, x) = \mathrm{softmax}(W h_t + b).
\]
\endgroup
Here, $f_{\theta}$ denotes the decoder network parameterized by $\theta$, while $W$ and $b$ are the output projection parameters mapping hidden states to vocabulary logits. Because representation directions in $h_t$ encode the intent-slot associations learned during training, suppressing $p_{\theta}(i_f \mid x)$ does not necessarily modify these directions. As shown in Table~\ref{tab:02}, intent-level suppression reduces intent accuracy but leaves conditional slot recoverability largely intact. Consequently, the conditional mapping $p_{\theta}(s \mid i_f, x)$ remains active, and correct slots can still be generated under forced-prefix decoding. We refer to this structural phenomenon as \emph{capability persistence}. Effective selective unlearning should therefore intervene in the representation space to weaken intent-conditioned slot generation rather than only suppress marginal intent prediction.
\section{BSU: Binding Subspace Unlearning}
Building on this formulation, we introduce \textit{Binding Subspace Unlearning (BSU)}, a two-stage unlearning framework that intervenes in the model's representation space to mitigate capability persistence. Rather than suppressing only the marginal intent probability $p_{\theta}(i_f \mid x)$, BSU targets hidden-state directions associated with intent-conditioned slot generation. Let $\theta_0$ be a pretrained SLU model and $\theta$ a trainable copy initialized from $\theta_0$. Given a target intent $I_F$, the objective is to reduce the conditional distribution $p_{\theta}(s \mid i_f, x)$ under forced-prefix decoding while preserving performance on retained intents. BSU achieves this through two stages: \textit{\textbf{(i)}} identifying representation directions enriched when the target intent $I_F$ is present, and \textit{\textbf{(ii)}} reducing model sensitivity along these directions to weaken the conditional mapping.

\vspace{-0.2em}
\subsection{Stage I: Binding Subspace Identification}
\label{sec:subspace_id}
\vspace{-0.2cm}
The goal of Stage I is to identify representation directions that are statistically enriched when the target intent $I_F$ is present. These directions approximate components of the hidden state that influence the conditional slot-generation distribution $p_\theta(s \mid i_f, x)$. Since next-token probabilities are computed from hidden states, analyzing hidden representations at slot positions allows us to probe directions associated with intent-conditioned slot generation.

\noindent\textit{\underline{(1) Hidden-State Extraction. }} Semantic outputs follow a structured prefix format consisting of an intent token followed by slot tokens. We estimate the boundary between intent and slot tokens and extract hidden states at slot positions using teacher-forced decoding \cite{williams1989learning}:
\begingroup
\setlength{\abovedisplayskip}{2pt}
\setlength{\belowdisplayskip}{2pt}
\setlength{\abovedisplayshortskip}{2pt}
\setlength{\belowdisplayshortskip}{2pt}
\[
h_t^{(\ell)} = f_{\theta}^{(\ell)}(x, y_{<t}),
\qquad
t > t_{\text{prefix}}.
\]
\endgroup
Here,  $h_t^{(\ell)}$ denotes the decoder representation at layer $\ell$. Teacher forcing ensures alignment between hidden states and ground truth slot structure, avoiding variability induced by decoding errors.

\vspace{0.3em}
\noindent\textit{\underline{(2) Covariance Contrast. }}Using these slot-position representations $h_t^{(\ell)}$, we compute empirical covariance matrices over hidden states for both the forget dataset $\mathcal{D}_F$ and the retain dataset $\mathcal{D}_R$ at each layer $\ell$. We then form the contrast matrix as:
\begingroup
\setlength{\abovedisplayskip}{4pt}
\setlength{\belowdisplayskip}{4pt}
\setlength{\abovedisplayshortskip}{4pt}
\setlength{\belowdisplayshortskip}{4pt}
\[
M^{(\ell)}
=
\mathrm{Cov}_{\mathcal{D}_F}^{(\ell)}
-
\mathrm{Cov}_{\mathcal{D}_R}^{(\ell)}
\]
\endgroup
This contrast highlights representation directions that have higher variance when the target intent $I_F$ is present relative to retained data. Since next-token probabilities are computed as affine transformations of hidden states, directions that consistently vary under $D_F$ are expected to contribute more strongly to the conditional slot likelihood $p_\theta(s \mid i_f, x)$.

\vspace{0.4em}
\noindent\textit{\underline{(3) Subspace Extraction. }}Finally, we compute the top positive eigenvectors of $M^{(\ell)}$, denoted as $U^{(\ell)} \in \mathbb{R}^{d \times k}$. These eigenvectors define a low-dimensional subspace that approximates representation directions associated with intent-conditioned slot generation. This construction is statistical rather than exact; it identifies directions enriched under $I_F$ without assuming strict disentanglement of representations.

\subsection{Stage II: Subspace-Guided Capability Attenuation}
\label{sec:subspace_att}
Stage~II attenuates intent-conditioned slot generation by reducing model sensitivity along the identified binding subspaces $U^{(\ell)}$. Rather than modifying only output distributions, BSU targets the first-order sensitivity of the conditional likelihood with respect to hidden states. For slot positions associated with $I_F$, we compute the gradient of the teacher-forced conditional log-likelihood:
\begingroup
\setlength{\abovedisplayskip}{2pt}
\setlength{\belowdisplayskip}{2pt}
\setlength{\abovedisplayshortskip}{2pt}
\setlength{\belowdisplayshortskip}{2pt}
\[
g_t^{(\ell)}=\nabla_{h_t^{(\ell)}}\log p_\theta(s\mid i_f,x)
\]
\endgroup
If slot generation under $I_F$ relies on particular representation directions, the gradient will exhibit large components along them. We therefore project the gradients onto the binding subspaces and penalize their squared magnitude:
\begingroup
\setlength{\abovedisplayskip}{2pt}
\setlength{\belowdisplayskip}{2pt}
\setlength{\abovedisplayshortskip}{2pt}
\setlength{\belowdisplayshortskip}{2pt}
\[
\mathcal{L}_{\mathrm{bind}}
=
\sum_{\ell=1}^{L}
\sum_{t>t_{\mathrm{prefix}}}
\left\lVert
U^{(\ell)}U^{(\ell)\top}g_t^{(\ell)}
\right\rVert_2^2 
\]
\endgroup
Minimizing $\mathcal{L}_{\mathrm{bind}}$ reduces sensitivity along representation directions enriched under $I_F$, weakening intent-conditioned slot generation even when the intent prefix is externally supplied. The attenuation is achieved through parameter updates and introduces no inference-time overhead.

\begingroup
\scriptsize
\setlength{\tabcolsep}{1.6pt}
\renewcommand{\arraystretch}{0.85}
\setlength{\extrarowheight}{0.8pt}
\begin{table*}[!t]
\centering
\vspace{-1cm}
\caption{Selective capability unlearning on \textbf{SLURP} and \textbf{SpeechMassive}. We report intent accuracy ($I_F$), slot macro-F1 ($F1_F$), BRR@10, and semantic similarity (Sim.) on the target $D_F$ and retain $D_R$ sets, respectively. \textbf{Bold}: Best, \underline{Underlined:} Second-Best.}
\label{tab:02}
\vspace{-0.3cm}
\begin{adjustbox}{max width=\textwidth}
\begin{tabular}{|l|l|cccc|cccc||cccc|cccc|}
\toprule
\multirow{3}{*}{\textbf{Dataset}}
& \multirow{3}{*}{\textbf{Method}}
& \multicolumn{8}{c||}{\textbf{NeMo-Conformer-Transformer}}
& \multicolumn{8}{c}{\textbf{NeMo-SSL-Conformer}} \\
\cmidrule(lr){3-10}\cmidrule(lr){11-18}
&
& \multicolumn{4}{c|}{\textbf{Forget Set ($D_F$)} }
& \multicolumn{4}{c||}{\textbf{Retain Set ($D_R$)}}
& \multicolumn{4}{c|}{\textbf{Forget Set ($D_F$)}}
& \multicolumn{4}{c}{\textbf{Retain Set ($D_R$)}} \\
\cmidrule(lr){3-6}\cmidrule(lr){7-10}\cmidrule(lr){11-14}\cmidrule(lr){15-18}
&
& $I_F\,\downarrow$ & $F1_F\,\downarrow$ & $\mathrm{BRR}@10\,\downarrow$ & $\mathrm{Sim.}\,\downarrow$
& $I_F\,\uparrow$ & $F1_F\,\uparrow$ & $\mathrm{BRR}@10\,\uparrow$ & $\mathrm{Sim.}\,\uparrow$
& $I_F\,\downarrow$ & $F1_F\,\downarrow$ & $\mathrm{BRR}@10\,\downarrow$ & $\mathrm{Sim.}\,\downarrow$
& $I_F\,\uparrow$ & $F1_F\,\uparrow$ & $\mathrm{BRR}@10\,\uparrow$ & $\mathrm{Sim.}\,\uparrow$ \\
\midrule

\multirow{8}{*}{\rotatebox[origin=c]{90}{\textbf{SLURP}}}
& \cellcolor{gray!23} Original
& \cellcolor{gray!23}95.27 & \cellcolor{gray!23}91.08 & \cellcolor{gray!23}92.64 & \cellcolor{gray!23}90.14
& \cellcolor{gray!23}89.75 & \cellcolor{gray!23}80.11 & \cellcolor{gray!23}91.17 & \cellcolor{gray!23}90.11
& \cellcolor{gray!23}91.05 & \cellcolor{gray!23}84.90 & \cellcolor{gray!23}84.42 & \cellcolor{gray!23}83.27
& \cellcolor{gray!23}89.16 & \cellcolor{gray!23}78.90 & \cellcolor{gray!23}85.17 & \cellcolor{gray!23}86.32 \\

& \cellcolor{yellow!20} Retrain
& \cellcolor{yellow!20}14.04 &\cellcolor{yellow!20} 20.13&\cellcolor{yellow!20} 18.59&\cellcolor{yellow!20}23.53
& \cellcolor{yellow!20}90.14 &   \cellcolor{yellow!20}86.58 &\cellcolor{yellow!20}81.38 &\cellcolor{yellow!20}82.10
& \cellcolor{yellow!20}13.72 & \cellcolor{yellow!20}16.58 &   \cellcolor{yellow!20}21.38 &   \cellcolor{yellow!20}22.19
& \cellcolor{yellow!20}90.15 &\cellcolor{yellow!20}80.43 & \cellcolor{yellow!20}81.70 &\cellcolor{yellow!20}83.27 \\

& GA      & \textbf{21.36} & \underline{31.78} & 91.14 & 86.92 & 54.15 & 52.09 & 87.62 & 83.40
& \underline{14.80} & \underline{22.50} & \underline{63.10} & \underline{66.40}
& 89.16 & 78.00 & 85.42 & 84.27 \\
& GA+GD   & 63.23 & 69.50 & 94.05 & 89.90 & 85.57 & 84.16 & 83.06 & 87.42
& 48.60 & 54.20 & 71.90 & 74.13
& 85.05 & 79.43 & 85.01 & 88.76 \\

& GA+KL   & 42.60 & 44.40 & 94.78 & 88.73 & 81.49 & 79.04 & 88.17 & 86.84
& 28.90 & 31.70 & 69.40 & 71.00
& 80.11 & 78.39 & 81.07 & 81.72 \\

& NPO     & 54.85 & 52.73 & 87.62 & 86.78 & 84.05 & 82.21 & 82.14 & 87.91
& 36.20 & 39.80 & 61.50 & 65.90
& 88.25 & 79.00 & 80.11 & 86.80 \\

& NPO+KL  & 84.17 & 82.62 & 84.05 & 82.79 & 89.68 & 82.94 & 85.07 & 80.17
& 62.70 & 58.10 & 79.81 & 76.20
& 89.01 & 80.51 & 83.30 & 89.05 \\
& RL  &58.7  & 63.41 &71.60  &80.32  &88.91  &81.88  &84.72  & 85.50
& 61.5 &66.28  & 75.30 & 81.00
& 90.01 &79.26  & 81.00 & 83.49 \\

\cline{2-18}
& \textcolor{gray}{\textbf{\textit{RS (Ablation)}}} & \textcolor{gray}{62.70} & \textcolor{gray}{71.59} & \textcolor{gray}{87.01} & \textcolor{gray}{88.0} & \textcolor{gray}{82.53} & \textcolor{gray}{82.01} & \textcolor{gray}{90.14} & \textcolor{gray}{89.01} & \textcolor{gray}{50.15} & \textcolor{gray}{50.59} & \textcolor{gray}{69.11} & \textcolor{gray}{70.16} & \textcolor{gray}{90.18} & \textcolor{gray}{73.87} & \textcolor{gray}{80.10} & \textcolor{gray}{83.15} \\

& \cellcolor{blue!12}\textbf{\textit{BSU (Ours)}}
& \cellcolor{blue!12}\underline{28.40} & \cellcolor{blue!12}\textbf{16.22} & \cellcolor{blue!12}\textbf{22.10} & \cellcolor{blue!12}\textbf{24.80}
& \cellcolor{blue!12}87.90 & \cellcolor{blue!12}83.62 & \cellcolor{blue!12}88.53 & \cellcolor{blue!12}90.8
& \cellcolor{blue!12}\textbf{11.20} & \cellcolor{blue!12}\textbf{14.10} & \cellcolor{blue!12}\textbf{16.30} & \cellcolor{blue!12}\textbf{18.70}
& \cellcolor{blue!12}87.40 & \cellcolor{blue!12}79.14 & \cellcolor{blue!12}83.90 & \cellcolor{blue!12}86.16\\
\midrule \midrule
\multirow{8}{*}{\rotatebox[origin=c]{90}{\shortstack{\textbf{SpeechMassive}}}}
& \cellcolor{gray!23} Original
& \cellcolor{gray!23}81.55  &\cellcolor{gray!23}54.90  & \cellcolor{gray!23}79.02   & \cellcolor{gray!23}76.39
& \cellcolor{gray!23} 79.54  &\cellcolor{gray!23} 52.60  & \cellcolor{gray!23}78.49   & \cellcolor{gray!23} 77.83
& \cellcolor{gray!23} 77.38  &\cellcolor{gray!23}52.03  & \cellcolor{gray!23}77.02  & \cellcolor{gray!23} 74.70
& \cellcolor{gray!23} 73.66 &\cellcolor{gray!23} 51.09 & \cellcolor{gray!23} 76.02  & \cellcolor{gray!23}74.11 \\

& \cellcolor{yellow!20} Retrain
& \cellcolor{yellow!20} 12.74 & \cellcolor{yellow!20}23.01 &\cellcolor{yellow!20}19.77 &\cellcolor{yellow!20}21.80
& \cellcolor{yellow!20}80.52 & \cellcolor{yellow!20}54.11 & \cellcolor{yellow!20}80.14 & \cellcolor{yellow!20}79.44
& \cellcolor{yellow!20} 10.95 &\cellcolor{yellow!20}20.14 &\cellcolor{yellow!20}22.57 &\cellcolor{yellow!20}21.87
& \cellcolor{yellow!20}74.66 & \cellcolor{yellow!20}51.63 & \cellcolor{yellow!20}77.83 & \cellcolor{yellow!20}76.81 \\
& GA      &33.15  &37.91  &31.77  &32.16  &48.66  &41.05  &45.92  &47.37  & 40.14 &45.80  &40.57  &44.60
&51.22  &46.03  &51.52  &50.87  \\
& GA+GD   &77.51  &54.14  &78.01  &76.05  &78.05  &50.11  &68.50  &70.59  &74.06  &50.17  &67.55  &67.99  
&72.19  &49.17  &68.05 &67.00  \\
& GA+KL   &67.30  &48.01  &68.22  &71.00  &75.90  &53.80  &70.11  &71.08  &72.20  &48.68  &74.05  &74.70  &73.22  & 51.33 &73.59  &74.60  \\
& NPO     &39.56  &44.17  &35.91  &35.07  &70.86  &51.73  &66.24  &67.38
&50.19  &50.11  &49.83  &49.17  &72.94  &53.16  &69.88  &70.42 \\
& NPO+KL  &55.14  &49.76  &56.87  &58.43  &74.12  &53.92  &70.58  &71.16
&59.73  &51.34  &60.27  &61.43  &75.02  &55.94  &71.46  &72.08 \\
& RL      &28.63  &36.42  &\textbf{24.94}  &\textbf{26.18}  &68.74  &50.91  &64.07  &65.32
&\underline{34.76}  &\underline{41.27}  &\underline{30.46} &\underline{32.11}  &70.38  &52.48  &66.29  &67.44 \\
\cline{2-18}
& \textcolor{gray}{\textbf{\textit{RS (Ablation)}}}
  & \textcolor{gray}{51.07} & \textcolor{gray}{47.50} & \textcolor{gray}{57.39} & \textcolor{gray}{59.47}
  & \textcolor{gray}{67.92} & \textcolor{gray}{49.88} & \textcolor{gray}{63.15} & \textcolor{gray}{64.29}
  & \textcolor{gray}{53.74} & \textcolor{gray}{45.03} & \textcolor{gray}{54.16} & \textcolor{gray}{51.97}
  & \textcolor{gray}{69.81} & \textcolor{gray}{51.44} & \textcolor{gray}{65.88} & \textcolor{gray}{66.73} \\

& \cellcolor{blue!12}\textbf{\textit{BSU (Ours)}}
  & \cellcolor{blue!12}\textbf{22.41}  & \cellcolor{blue!12}\textbf{24.33 } & \cellcolor{blue!12}\underline{29.87}  & \cellcolor{blue!12}\underline{29.94}
  & \cellcolor{blue!12}79.14  & \cellcolor{blue!12}56.47  & \cellcolor{blue!12}78.36  & \cellcolor{blue!12}77.53
  & \cellcolor{blue!12}\textbf{25.37}  & \cellcolor{blue!12}\textbf{27.48}  & \cellcolor{blue!12}\textbf{23.64}  & \cellcolor{blue!12}\textbf{26.17}
  & \cellcolor{blue!12}82.73  & \cellcolor{blue!12}58.26  & \cellcolor{blue!12}80.14  & \cellcolor{blue!12}79.47 \\
\bottomrule

\end{tabular}
\end{adjustbox}
\vspace{-2em}
\end{table*}
\endgroup
\subsection{Overall Unlearning Objective}
\label{sec:overall_obj}
\vspace{-0.15cm}
For the final update step, we build on the standard forget--retain fine-tuning objective \cite{ren2025sok,kurmanji2023towards}. Let $L_F=-\mathbb{E}_{(x,y)\sim D_F}\log p_\theta(y\mid x)$ and $L_R=-\mathbb{E}_{(x,y)\sim D_R}\log p_\theta(y\mid x)$ denote the negative log-likelihood losses on the forget and retain sets, respectively. Since minimizing $L_F$ would further fit the forget examples, we reverse its sign in the unlearning objective. Thus, gradient descent on the total objective performs gradient ascent on $L_F$, suppressing the target capability, while $L_R$ preserves performance on non-target intents:
\[
\small
L_{\mathrm{base}}=-L_F+\lambda_{\mathrm{ret}}L_R .
\]
To stabilize optimization and limit deviation from the original model $\theta_0$, we introduce a retain-set KL regularizer $L_{\mathrm{kl}}$, which keeps the updated model close to the original distribution on $D_R$. We then incorporate the binding loss $L_{\mathrm{bind}}$ from Section~3.2 to reduce sensitivity along intent-conditioned slot-generation directions. The final objective is
\[
\small
L=-L_F+\lambda_{\mathrm{ret}}L_R+\lambda_{\mathrm{kl}}L_{\mathrm{kl}}+\lambda_{\mathrm{bind}}L_{\mathrm{bind}} .
\]
All terms are optimized with standard gradient descent; the negative coefficient on $L_F$ induces ascent on the forget-set NLL, while the retain, KL, and binding terms are minimized normally.

\section{Experimental Setup}
\subsection{ Datasets and Task Setup}
\noindent \textbf{Datasets. }We evaluate selective capability unlearning on SLURP \cite{bastianelli2020slurp}, a standard end-to-end SLU benchmark dataset with each utterance annotated by an intent and corresponding slot-value pairs. To assess cross-lingual robustness, we additionally evaluate the French subset of SpeechMASSIVE \cite{lee2024speech}, which follows the same semantic mapping. For each dataset, we select a target intent $I_F$ and partition the data into a forget set $D_F$ (utterances labeled with $I_F$) and a retain set $D_R$ (remaining intents).

\vspace{0.3em}
\noindent \textbf{Implementation Details. }All experiments use an end-to-end SLU architecture consisting of a Conformer acoustic encoder and a Transformer-based semantic decoder. We evaluate two model variants that share the same architecture, tokenizer, decoder, and training protocol, differing only in encoder initialization. In the \textit{(i) ASR-initialized SLU} setting, the encoder is initialized from a supervised ASR checkpoint (Conformer--Transformer--Large, NeMo ASR-Set~3.0). In the \textit{(ii) SSL-initialized SLU} setting, the encoder is instead initialized from a self-supervised speech representation while keeping all other components identical. Unless otherwise stated, we use $\lambda_{\mathrm{ret}}=1.0$, $\lambda_{\mathrm{kl}}=0.1$, and $\lambda_{\mathrm{bind}}=0.5$.

\subsection{Evaluation Metrics}
\vspace{-0.1cm}
We evaluate selective capability unlearning using both surface-level and behavioral metrics. Surface metrics include intent accuracy and slot F1 following the SLURP protocol \cite{bastianelli2020slurp}. To measure residual capability beyond exact prediction, we adopt top-k decoding and embedding-based similarity, widely used metrics in generative unlearning \cite{chen2021evaluating,zhang2019bertscore,yao2024large,carlini2021extracting,holtzman2019curious}.

\noindent \textbf{BRR@10 (Beam Retrieval Rate \cite{vijayakumar2018diverse}) }
BRR@10 measures whether the forgotten capability remains recoverable under beam search.
For each utterance $x \in \mathcal{D}_F$, we decode conditioned on the ground truth
intent prefix $I_F$. Let $y$ be the ground truth semantic frame and
$\hat{y}^{(k)}$ the $k$-th beam hypothesis with beam size $K=10$. We define:
\begingroup
\setlength{\abovedisplayskip}{1pt}
\setlength{\belowdisplayskip}{2pt}
\setlength{\abovedisplayshortskip}{1pt}
\setlength{\belowdisplayshortskip}{2pt}
\[
\mathrm{BRR@10}
=
\frac{1}{|\mathcal{D}_F|}
\sum_{(x,y)\in \mathcal{D}_F}
\mathbf{1}
\left(
\exists\, k \le K :
\hat{y}^{(k)} = y
\right)
\]
\endgroup

\vspace{-0.1cm}
\noindent \textbf{Semantic Similarity \cite{reimers2019sentence} }To capture residual behavior beyond exact matching, we compute cosine similarity between embeddings $E(\hat{y})$ and $E(y)$, where $E(\cdot)$ maps structured semantic frames to vector representations. The cosine similarity is defined as follows:
\vspace{-0.1cm}
\begingroup
\setlength{\abovedisplayskip}{1pt}
\setlength{\belowdisplayskip}{2pt}
\setlength{\abovedisplayshortskip}{1pt}
\setlength{\belowdisplayshortskip}{2pt}
\[
{\tiny
\mathrm{Sim}(\hat{y}, y)
=
\frac{E(\hat{y})^\top E(y)}
{\lVert E(\hat{y}) \rVert_2 \, \lVert E(y) \rVert_2}
}
\]
\endgroup
A higher $\mathrm{BRR}@10$ indicates recoverable capability under beam search, while higher semantic similarity reflects stronger alignment with the ground-truth frame despite surface differences.


\subsection{Baselines }
\vspace{-0.1cm}
As no prior work explicitly studies capability unlearning in end-to-end SLU, we adapt canonical machine unlearning methods for our setting. We evaluate the Gradient Ascent (GA) family, which maximizes the loss on the forget set to disrupt target representations \cite{golatkar2020eternal,yao2024large}, along with two stabilized variants: GA+GD, which additionally trains on retain data, and GA+KL, which applies KL regularization to the original model. We further include Negative Preference Optimization (NPO) and NPO+KL, which penalizes distribution alignment with forget-set outputs via a preference objective, and Random Label (RLabel), which replaces forget-set labels with random targets to simulate capability removal \cite{graves2021amnesiac}.
\begin{figure}[t!]
\centering
\begin{subfigure}[t]{0.46\columnwidth}
  \centering
   \includegraphics[height=0.78\columnwidth, trim=10pt 0pt 0pt 0pt, clip]{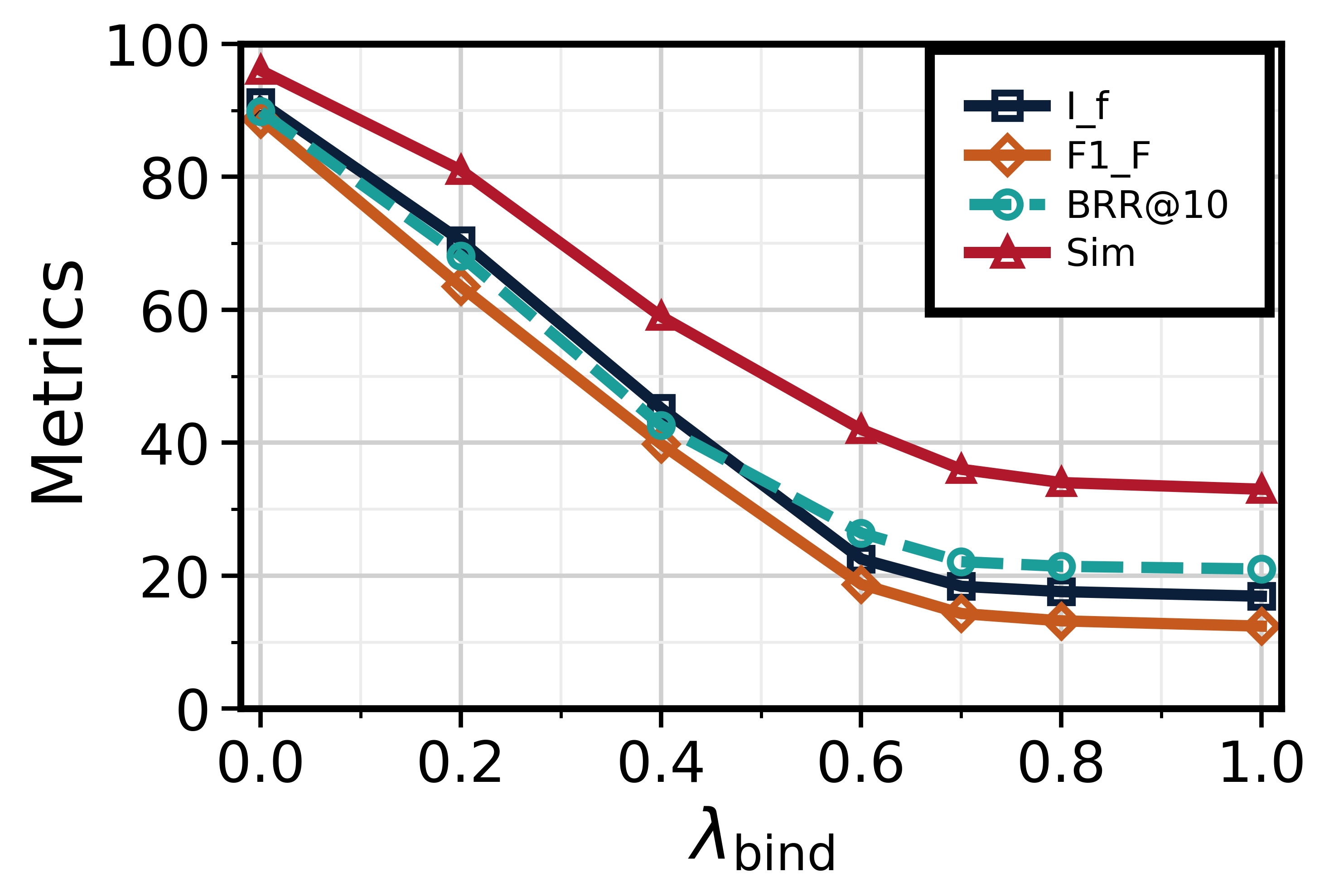}
   \vspace{-0.6cm}
  \caption{Effect on Forget Set}
  \label{fig:twoinrow:a}
\end{subfigure}\hfill
\begin{subfigure}[t]{0.46\columnwidth}
  \centering
  \includegraphics[height=0.78\columnwidth, trim=39pt 0pt 0pt 0pt, clip]{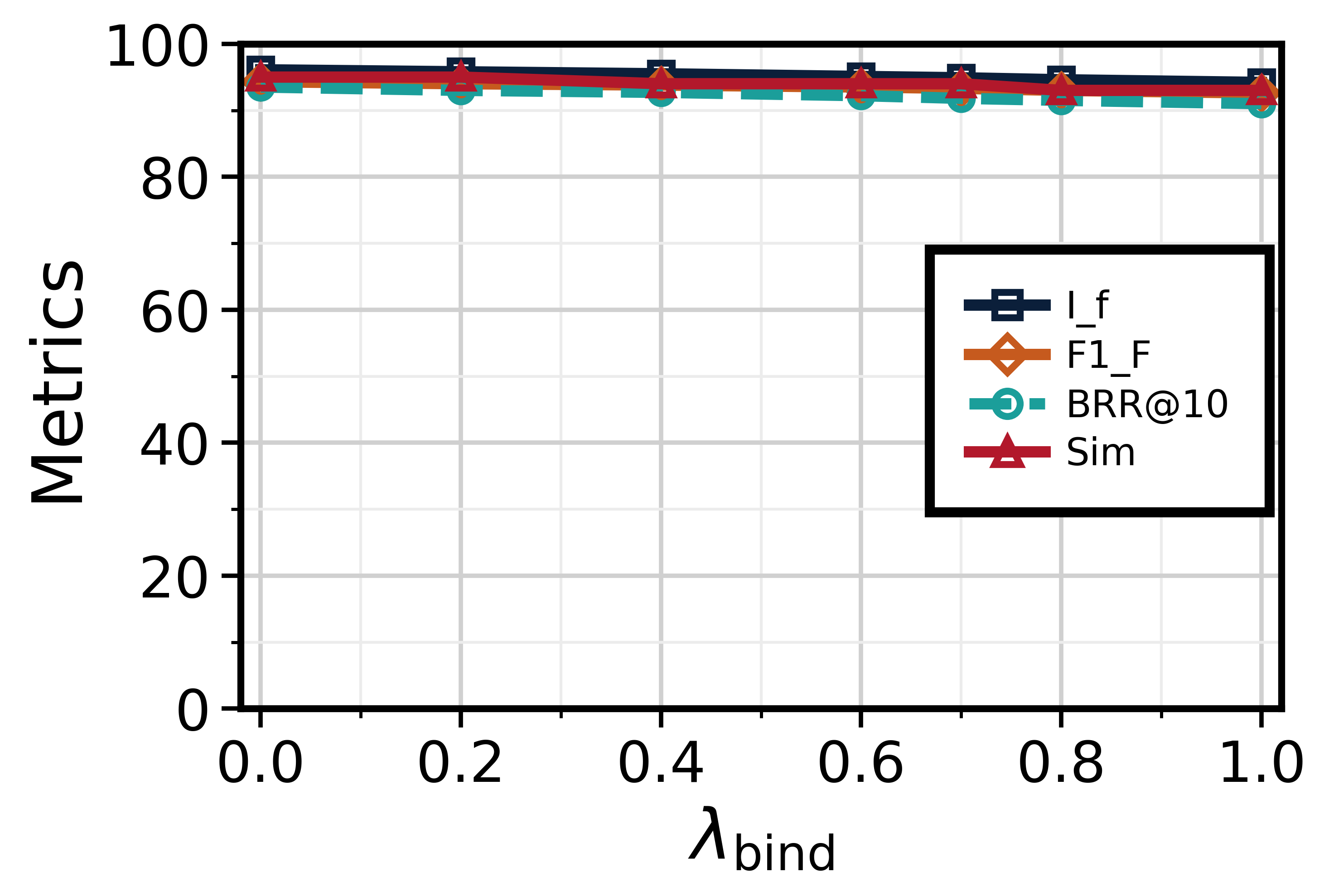}
     \vspace{-0.6cm}
    \caption{Effect on Retain Set}
  \label{fig:twoinrow:b}
\end{subfigure}
   \vspace{-0.25cm}
\caption{\textbf{Effect of the binding regularizer $\lambda_{bind}$}. Increasing $\lambda_{bind}$ suppresses the target capability on the forget set ($D_F$), reducing all metrics, while performance on the retain set ($D_R$) remains largely stable.}
\label{fig:twoinrow}
\end{figure}
\begin{figure}[t]
\centering
\includegraphics[width=0.85\columnwidth]{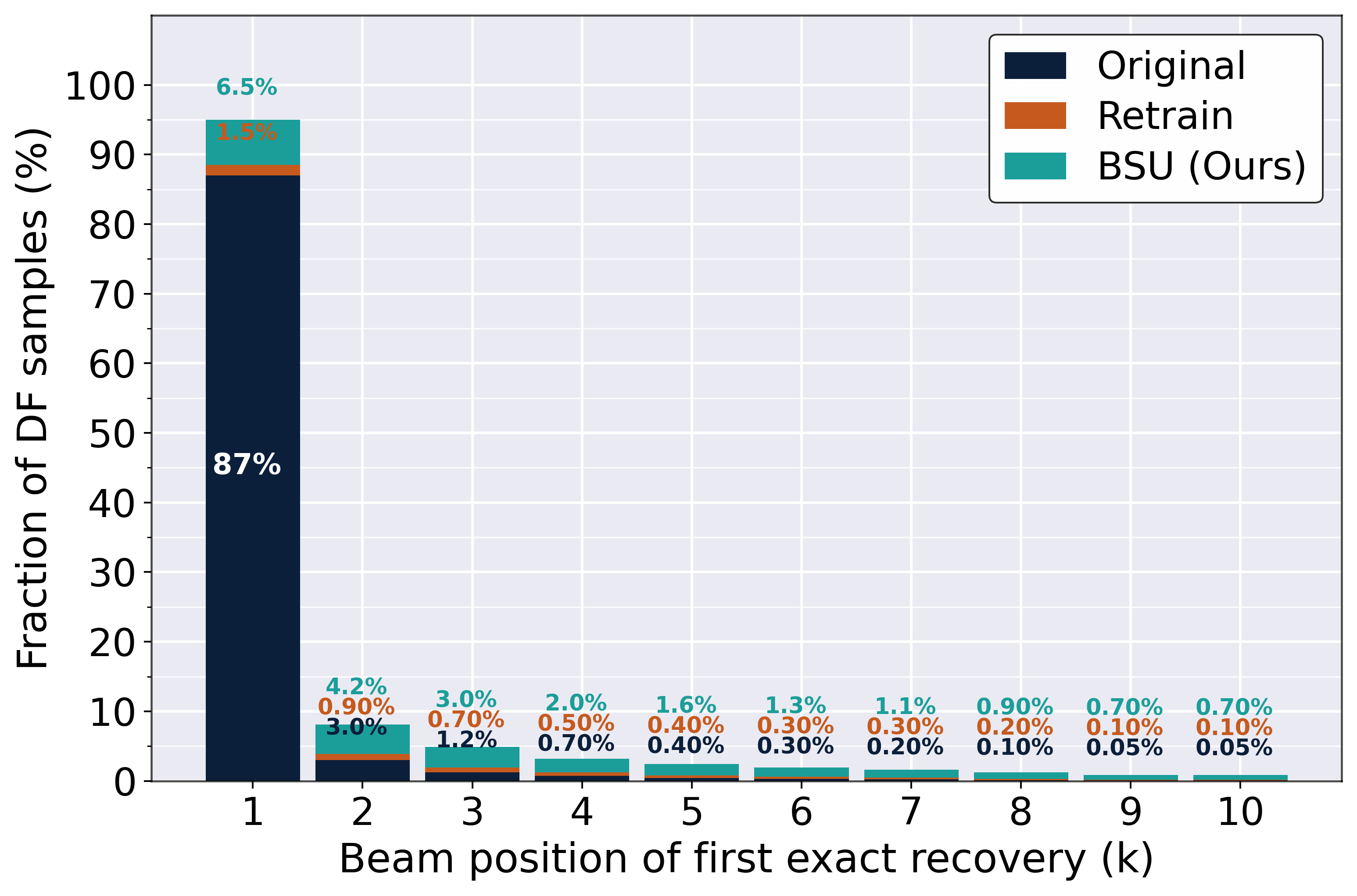}
\vspace{-0.3cm}
\caption{\textbf{Exact recovery across beam positions.} We report the Beam Retrieval Rate (BRR@10), evaluating whether the correct semantic frame appears within the top-K beam hypotheses. Higher values indicate stronger target recoverability.}
\label{fig:beam_exact_recovery}
\end{figure}
\vspace{-0.35cm}
\section{Results and Analysis}
\subsection{Comparison with Baselines}
\vspace{-0.2cm}
We evaluate whether slot content associated with the target intent can still be generated when the intent prefix is provided at test time, while the performance on non-target intents is preserved. As shown in Table~\ref{tab:02}, gradient and preference-based baselines (GA, GA+KL, NPO) reduce marginal intent prediction on the forget set ($D_F$) in several settings. However, their forced-prefix recovery scores often remain high, indicating that intent-conditioned slot content can still be recovered under forced-prefix decoding. In contrast, \textit{\textbf{BSU (ours)}} leads to pronounced reductions across all forget-set metrics. For SLURP with NeMo Conformer-Transformer, BRR@10 decreases from 92.64 to 22.10 and semantic similarity decreases from 90.14 to 24.80. Comparable reductions are observed with the SSL-initialized Conformer, where BRR@10 decreases from 84.42 to 16.30 and semantic similarity decreases from 83.27 to 18.70. The same trend is observed on SpeechMassive, where BSU also reduces BRR@10 and semantic similarity across both model initializations. At the same time, retain-set performance remains stable relative to the original and retrain baselines in most settings, indicating selective unlearning.

\subsection{Residual Capability Analysis} 
\noindent \textbf{Random Space Analysis. }We include a Random Space (RS) ablation in Table~\ref{tab:02} to test whether unstructured perturbations in the representation space can induce forgetting. RS perturbs representations along random directions that are not aligned with the subspace encoding the target intent–slot dependency. Consequently, the perturbation does not systematically disrupt the underlying capability, resulting in limited suppression.

\vspace{0.3em}
\noindent \textbf{Binding Regularizer Sensitivity. }Figure~\ref{fig:twoinrow} analyzes as $\lambda_{bind}$ increases, performance on the forget split ($I_F$, $F1_F$, BRR@10) decreases steadily, indicating progressively stronger suppression of the target capability. In contrast, metrics on the retain split remain largely stable across the same range of $\lambda_{bind}$.

\vspace{0.3em}
\noindent \textbf{Residual Capability Recovery. }Figure~\ref{fig:beam_exact_recovery} demonstrates whether the target capability can be regenerated during decoding by leveraging the beam position of the first exact semantic-frame recovery. The stacked bars show the fraction of forget-set ($D_F$) samples whose correct frame first appears at beam position $k$ ($1 \leq k \leq 10$). The original model recovers the correct frame for a large portion of samples at early beam positions, indicating strong memorization of the target behavior. Both retraining and BSU significantly lower recovery across all beam positions, indicating that exact recovery is strongly reduced, even along alternative decoding paths.
\vspace{-0.2cm}
\section{Conclusion}
\vspace{-0.2cm}
In this work, we show that suppressing marginal intent prediction alone does not eliminate the conditional mapping governing slot generation, leading to capability persistence under forced-prefix decoding. We propose \textit{Binding Subspace Unlearning (BSU)}, which removes this dependency by targeting representation-level binding directions. Experiments show substantial reductions in recoverable slot behavior while preserving retained-intent performance. Future work includes extending this approach to broader generative tasks and developing principled capability-level unlearning methods for privacy-centric applications.

\section{Acknowledgments}
We acknowledge the institutional and computational support provided by the Department of Data Science and Engineering, Indian Institute of Science Education and Research Bhopal.

\section{Generative AI Use Disclosure}
Generative AI tools were used only for language editing and polishing. All scientific content, experimental design, analyses, results, and conclusions were developed and verified by the authors, who take full responsibility for the content of this paper.

\bibliographystyle{IEEEtran}
\bibliography{mybib}

\end{document}